\documentclass[runningheads]{llncs}

\usepackage{graphicx}
\usepackage{microtype}
\usepackage[caption=false]{subfig} 

\usepackage{makecell}  
\usepackage{xcolor,colortbl}

\definecolor{Gray}{gray}{0.85}
\definecolor{LightCyan}{rgb}{0.88,1,1}
\newcolumntype{a}{>{\columncolor{Gray}}l}
\newcolumntype{b}{>{\columncolor{LightCyan}}l}

\newcolumntype{T}{>{\tiny}c} 

\usepackage{booktabs} 

\usepackage{amsmath}
\usepackage{microtype}
\usepackage{multirow}

\usepackage[acronym]{glossaries}
\newacronym{TOPIC}{TOPIC}{Empirical study of Machine Learning Classifier Evaluation Metrics behavior in Massively Imbalanced and Noisy data}

\begin{document}

\title{Empirical study of Machine Learning Classifier Evaluation Metrics behavior in Massively Imbalanced and Noisy data}
\titlerunning{Empirical study : Metrics in Massively Imbalanced and Noisy data sets}

\author{Gayan K. Kulatilleke\inst{1} \and Sugandika Samarakoon\inst{2}}
\authorrunning{GK. Kulatilleke \and S. Samarakoon}
%
\institute{
Work outlined in this paper is part of the author’s MSc dissertation at Queen Mary University of London, 2017.\\\email{tidalbobo@gmail.com}
\and
\email{sugandikasamarakoon@gmail.com}
}
%
\maketitle  
\begin{abstract}
With growing credit card transaction volumes, the fraud percentages are also rising, including overhead costs for institutions to combat and compensate victims.
The use of machine learning into the financial sector permits more effective protection against fraud and other economic crime. Suitably trained machine learning classifiers help proactive fraud detection, improving stakeholder trust and robustness against illicit transactions.
However, the design of machine learning based fraud detection algorithms has been challenging and slow due the massively unbalanced nature of fraud data and the challenges of identifying the frauds accurately and completely to create a gold standard ground truth.
Furthermore, there are no benchmarks or standard classifier evaluation metrics to measure and identify better performing classifiers, thus keeping researchers in the dark. 

In this work, we develop a theoretical foundation to model human annotation errors and extreme imbalance typical in real world fraud detection data sets. By conducting empirical experiments on a hypothetical classifier, with a synthetic data distribution approximated to a popular real world credit card fraud data set, we simulate human annotation errors and extreme imbalance to observe the behavior of popular machine learning classifier evaluation matrices. We demonstrate that a combined F1 score and g-mean, in that specific order, is the best evaluation metric for typical imbalanced fraud detection model classification.

\end{abstract}

\section{Introduction} \label{Introduction}
Credit cards play a major role in modern finance \cite{zojaji2016survey} with its ease, popularity, ubiquity, cross boarder reach and instantaneous confirmation. Unfortunately, this has also resulted in a fertile ground for fraud.
According to Financial Fraud Action UK, in 2016, the credit card transaction volume was 19.1 billion. This is an increase of 10\% from previous year. Fraud losses amounted to £618.0 million, a 9\% increase from the previous year, and the fifth consecutive year of increases. Unfortunately, the current prevention rates are only 0.6/£.

Credit card frauds impacts all stakeholders, with the ultimately burden on the general public, who has to pay increased bank charges and fees to buffer merchant's fraud costs \cite{kulatilleke2022Challenges}. Loss of trust by users and reputation loss of the merchants have long term implications on the financial system, including economic inefficiencies such as higher risk provisions, slowdown, and depression \cite{kulatilleke2022Challenges}. US LexisNexis Group \footnote{Lexisnexis.com} reported that, for each \$1 of fraud, the additional costs related to card and merchandise replacement, charge-backs and administrative overheads results in a true loss of \$2.40 per \$ in 2016, which is an 8\% increase from the previous year. Thus, fraud detection systems are essential to minimize and counter increasing frauds and fraud costs \cite{zareapoor2015application}. 

Traditionally, fraud classification was handled using data mining \cite{duman2013solving}. However, the massive data imbalance in credit card fraud data results in very low number of frauds (positives), which means that a false-negative (not detecting a fraud) has very serious consequences. \cite{duman2013solving} argues that fraudulent transactions of larger values or higher card limits are more important from a detection perspective and the lack of arbitrary costs is a main known deficiency of data mining approaches. 

Credit Card Fraud Detection can be addressed via machine learning \cite{kulatilleke2022Credit}. It provides a means to study vast amounts of data and discover hidden and also evolving patterns of malicious occurrences \cite{sathyapriya2017big}. Using a labelled set of sample transactions, machine learning has the ability to "learn" how to detect and flag frauds in new transactions \cite{kulatilleke2022Challenges,ebert2016machine}. Thus, use of machine learning into the financial sector presents a promising and efficient context for fraud detection \cite{makki2017fraud}. 

Automated detection can handle larger quantities of transaction data, with a lower cost in a shorter period of time with consistency. By identifying a smaller subset of true positives, i.e., detecting fraudulent transactions as frauds correctly, human reviewers can focus and prioritize on more serious frauds and crimes with less manpower.

Unfortunately, credit card fraud detection via machine learning has several major constraints and challenges \cite{duman2013solving,zojaji2016survey}, including  massively unbalanced nature of fraud data, difficulty of identifying the frauds accurately and completely to create a gold standard ground truth and the lack of standard classifier evaluation metrics to measure and identify better performing \cite{kulatilleke2022Challenges}. 

Typical fraud data sets are massively imbalanced, typically bellow 1\% of fraud \cite{kulatilleke2022Challenges}. This unbalance has serious ramifications in machine learning. It is challenging to determine a proper evaluation metrics that can effectively score and select the best performing classifier. Traditional measures such as Accuracy cannot be used to identify better performing classifiers due to the massive imbalance in fraud data \cite{barandela2003strategies}. 
In the absence of an ideal acceptable evaluation metric \cite{zojaji2016survey}, or measurement score, it becomes impossible and misleading to attempt to determine the best model or the effects of any enhancements. Furthermore, comparative studies of evaluation metrics behavior under extreme imbalances is lacking. There is also a lack of studies focusing on the effect of human annotation errors on the final classifier score, from the perspective of the metrics measurement dynamics, when using different evaluation metrics.  

In this work we:
\begin{itemize}
    \item Derive a theoretical model to combine the human annotation error and machine classification error
    \item conduct experiments to observe behavior of popular evaluation metrics under diverse minority and annotation error compositions using a synthetic data set on a hypothetical model
    \item determine viable evaluation metrics appropriate for the evaluate the algorithmic performance of characteristic massively unbalanced (fraud) data sets
\end{itemize}

\section{Background}
\subsection{Characteristics of fraud data - the Massive imbalance}
Massive imbalance and wide variation in fraud percentages is characteristic of fraud detection data sets \cite{kulatilleke2022Challenges}.
Next we present two real world data sets to illustrate these characteristics. 

\begin{table}[t]
    \begin{tabular}{p{0.20\linewidth} p{0.20\linewidth} p{0.40\linewidth} p{0.20\linewidth}}
    \toprule
    Normal  & Frauds & Features            & Instances \\ \midrule
    284,315 & 492    & 30 (28 PCA + 2 Raw) & 284, 807  \\ \bottomrule
    \end{tabular}
    \caption{Primary PCA encoded (Pozzolo) Data set }
    \label{TABLE_PCA_primary_dataset}
\end{table}
Table~\ref{TABLE_PCA_primary_dataset} summarizes the features of an encoded real-world credit card transaction data set, comprising of 2 days in September 2013 by European cardholders \cite{dal2015calibrating}. The fraud percentage is 0.172\%, indicating massively imbalance.  
\begin{table}[t]
    \begin{tabular}{p{0.20\linewidth} p{0.20\linewidth} p{0.40\linewidth} p{0.20\linewidth}}
    \toprule
    Normal  & Frauds & Features  & Instances \\ \midrule
    23,364  & 6634   & 23 Raw   & 30,000   \\ \bottomrule
    \end{tabular}
    \caption{Secondary Data set }
    \label{TABLE_secondary_data set}
\end{table}
Table~\ref{TABLE_secondary_data set} shows another real-world data set consisting of raw (not encoded) features related to client information, payment history, status and credit limits of credit card customers in Taiwan from April 2005 to September 2005 \cite{lichman2013uci}. Its frauds are unbalanced with a percentage of 22\%, and significantly less imbalanced than former data set. 

According to \cite{kulatilleke2022Challenges} there are wide variations fraud detection data set in terms of feature definitions, size, feature size and fraud ratios; however, they are all massively imbalanced.


\subsection{Binary classification}
In machine learning, there are many evaluation metrics that measures very different perspectives of effectiveness \cite{zojaji2016survey}. Overall accuracy, or correctness cannot be used in typical massively unbalanced credit card transactions due to the high TN affecting the score. Thus, a higher accuracy score will not result in a necessarily good classifier \cite{kulatilleke2022Challenges}. In order to determine the goodness (indicated by a higher score), it is required consider the correct \textit{and} incorrect predictions as well as the relative weights of the classes. 
For a binary classification (i.e., differentiating between fraud and normal) problem, the confusion matrix shows all the four possible combinations and summarizes the performance of a classification algorithm. The number of correct and incorrect predictions are summarized with count values and broken down by each class, showing the mistakes made by the model for the frauds as well as the normals.

\begin{figure}
    \centering \includegraphics[width=0.5\columnwidth]{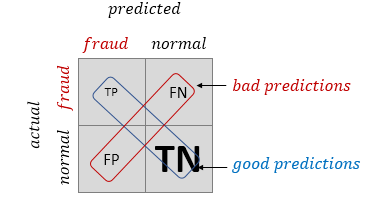}
    \caption{Confusion Matrix and terminology}
    \label{fig_confusionMatrix}
\end{figure}
Figure~\ref{fig_confusionMatrix} shows the confusion matrix for a credit card fraud classification problem. The TN, comprising normals form over 99.8\% (in the primary data set) against the minor amount of frauds of 0.2\%, which are shown in larger and smaller fonts to put them in perspective. The following definitions are used in the rest of this paper:
\begin{itemize}
    \item TP: true positives = the number of frauds detected correctly
    \item TN: true negatives = the number of normals detected correctly
    \item FP: false positives = the number of normals detected as frauds ((Type I errors)
    \item FN: false negatives = the number of frauds detected as normals, the number of frauds missed (Type II errors)
\end{itemize}
Using the above in various combinations, classifier performance can be measured in many ways \cite{west2016some,dal2017credit}. Table~\ref{TABLE_Matrices} lists some popular performance metrics for credit card fraud detection measurement using a binary classifier.

\begin{table}[t]
    \scriptsize
    \begin{tabular}{p{0.15\linewidth} p{0.40\linewidth} T}
    \toprule
    Metric  & Description   &  \\ \midrule
    \begin{tabular}{l}Accuracy\\Detection rate\end{tabular} & Overall correctness of the classifier  &  $\dfrac{TP+TN}{TP+TN+FP+FN}$       \\ \midrule
    \begin{tabular}{l}Precision\\Hit rate\end{tabular} & When a fraud is detected, how often was the detection correct? i.e.: TP/predicted yes  &   $\dfrac{TP}{TP+FP}$     \\ \midrule
    \begin{tabular}{l}Recall\\TPR\\Sensitivity\end{tabular} & When there was a fraud, how often did the classifier manage to  detect it? i.e.: TP/actual yes &  $\dfrac{TP}{TP+FN}$ \\ \midrule
    FPR  & Ratio of fraud detected incorrectly i.e.: FP/actual no &    $\dfrac{FP}{FP+TN}$      \\ \midrule
    \begin{tabular}{l}F1\\F-score\\F-measure\end{tabular} & Harmonic mean of precision and recall   &    $\dfrac{2\cdot sensitivity\cdot precision}{sensitivity+precision}$      \\ \midrule
    G-mean  & balance between classification performances on both the majority and minority classes   &   $\sqrt{sensitivity\cdot specificity}$       \\ \midrule
    AuROC  & Area under the ROC (graph of all possible true and false positive hit rates) &    $\int ROC$      \\ \midrule
    Cohen Kappa  & Score of the agreement between the prediction and the actual  &   $\dfrac{total\ accuracy-random\ accuracy}{1-random\ accuracy}$       \\ \midrule
    Mathew & Score of the correlation between the prediction and the actual       &   $\dfrac{TP\cdot TN-FP\cdot FN}{ \sqrt{(TP\cdot FP)(TP\cdot FN)(TN\cdot FP)(TN\cdot FN)}}$      \\ \bottomrule
    \end{tabular}
    \caption{Measurement Metrics}
    \label{TABLE_Matrices}
\end{table}

TN rate measures how well a classifier can recognize normal transactions \cite{zojaji2016survey}. However, in unbalanced data sets, TPR, TNR and average accuracy are misleading assessment measures \cite{dal2015calibrating}. While certain algorithms have lower accuracy, their relatively high sensitivity may result in better performance when considering cost minimization \cite{west2016some}.

Precision is not affected by large normals (negative classes), as it is more focused on the fraud (positive class), since it looks at TP and predicted positives. However, due this very reason, it misses on the FN and allows frauds to pass through, which is undesirable.

While generally a high confidence of TP (high Precision) and high detection rate of the positives (high Recall) is required, Precision comes at the cost of low Recall and vice versa, indicating an inverse relationship. The F1 score combines both Precision and Recall \cite{dal2015calibrating}, and computes the harmonic mean. Contrastingly \cite{west2016some} states that precision alone, with its highest variance between performance metrics, is best for assessing financial fraud detection solutions.

Notably, \cite{dal2015calibrating} recommends F1 score and g-mean as effective performance measures for unbalanced data sets. According to \cite{dal2017credit}, the typical performance measure for credit card fraud-detection is ROC curve (AuROC reduces the area of the ROC curve to a single comparable value), which is a global ranking measure while Average Precision (which corresponds to the area under the precision-recall curve) is also frequently used. 

According to \cite{sahin2013cost}, standard metrics such as accuracy, TPR, AuROC fail to evaluate performance; they suggested "Saved Loss Rate", the saved percentage of the potential loss, as a better metric.

\section{Methodology}
\subsection{The learning problem in the presence of annotation errors}
\begin{figure}
    \centering \includegraphics[width=0.8\columnwidth]{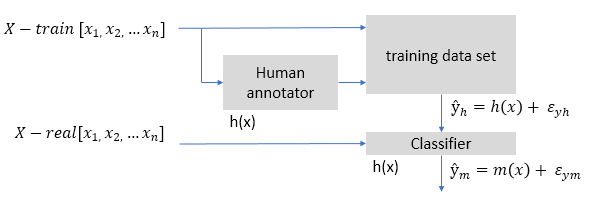}
    \caption{Formalizing the classification problem with real-world data, reproduced from \cite{kulatilleke2022Challenges}}
    \label{fig_learningProblem}
\end{figure}

We follow the same notation and reuse Figure~\ref{fig_learningProblem} from \cite{kulatilleke2022Challenges}; let $X$ be the feature vector of a card transaction. Further, $\textit{X-train}[x_1, x_2, \dots x_n]$ and $\textit{X-real}[x_1, x_2, \dots x_n]$ are respectively the train and test data sets. We use $h$ to denote human related operations and $m$ for machine related operations. Thus, $y, \hat{y}_h, \hat{y}_m$ are the real, human annotated and machine annotated labels and $\epsilon_{yh},\epsilon_{ym}$ are the human and machine error terms.

Importantly $\hat{y}_m$ can be considered a latent value. It is not possible to determine its accuracy comprehensively and completely, nor its accuracy distribution. 

Using the definitions, follwing \cite{kulatilleke2022Challenges}, we have:
\begin{equation} 
\begin{aligned}
    \text{human annotation:} \quad 	& \hat{y}_h=h(x)+ \epsilon_yh \\
    \text{machine classification:} \quad    & \hat{y}_m=m(x)+ \epsilon_ym \\
\end{aligned}
\label{eq-1}
\end{equation} 

As the machine (classifier) learns using annotated labels based on $\hat{y}_h$:
\begin{equation} 
    m = f(h, \epsilon_{yh})  
\label{eq-2}
\end{equation}

Thus, for \textit{any} given evaluation matrix of $S$:
\begin{equation} 
\begin{aligned}
    \text{Model error:} \quad 	& E_m = S( \hat{y}_h, \hat{y}_m) \\
    \text{Real error (model+human):} \quad   & E_r = S( y, \hat{y}_m) \\
\end{aligned}
\label{eq-3}
\end{equation} 

As the best model is obtained by minimizing the model error:
\begin{equation} 
    E_m^* = \text{arg min}(E_m)  
\label{eq-4}
\end{equation}

And the real world error is:
\begin{equation} 
    E_r^* = \text{arg min}(E_r)  
\label{eq-5}
\end{equation}

Essentially, in order to minimize the real error, \textit{both} components, human error and the model error needs to be minimized.
According to \cite{kulatilleke2022Challenges}, use of techniques to address the errors caused during the annotation stage will minimize $E_r$ making the classifier’s learning task easier and improve detection rates.
Also, an error in the classification of any of the minority classes (false negatives, i.e.: misclassification of fraud as normal) effects the model performance more than a more frequent majority class, as the error percentage (of a misclassified minority transaction) is high. Thus, the level of imbalance has an effect on error rates as it increases the impact of a single error, and hence its error percentage.  

\subsection{Empirical model formulation for simulating imbalance and noise}

\begin{figure}
    \centering \includegraphics[width=0.8\columnwidth]{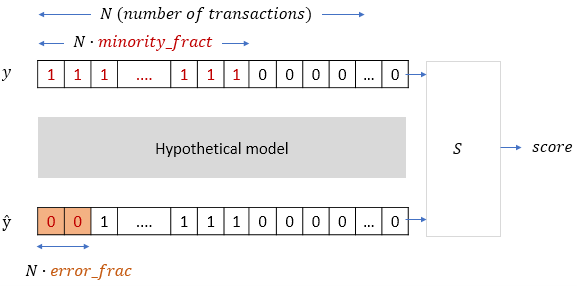}
    \caption{Logical diagram of the hypothetical model (M) to test Evaluation metrics (S)}
    \label{fig_HypotheticalModel}
\end{figure}
This section describes the mechanism used to investigate the evaluation metrics to determine an appropriate $S$, the optimal metrics, as shown in  Figure~\ref{fig_HypotheticalModel} and our assumptions. The proposed mechanism is independent of any model $M$, so that the behaviors of the metrics are universal and independent of any underlying model dynamics, making them usable in a broader range of use. Next, we explain how the effects of imbalance and human annotation error on evaluation metrics are investigated.

First, a synthetic label set $y$ is created using a pseudo-random generator. This is sent though a hypothetical model $M$. With no human errors $y = \hat{y}$ and performance of $M$ should be $1.00$. Next, we introduce human annotation errors of $error\_fract$ by inverting a label. Note that we are considering only binary classification in this work, though the approach can be extended to multi-class classification with ease. Thus with $error\_fract > 0$, $y \ne \hat{y}$, resulting in performance $< 1$. This performance is a function of how the metric reacts to noise and minority fraction.

As an illustrative example, with $N = 100$ and $minority\_fraction = 0.4$, the fraud and normals would be 40 and 60 respectively. If the $error\_rate = 2\%$, as shown in Figure~\ref{fig_HypotheticalModel}, $\hat{y}$ will have 2 values inverted to simulate human annotation errors affecting $M$. 

\begin{figure}
    \centering \includegraphics[width=1.0\columnwidth]{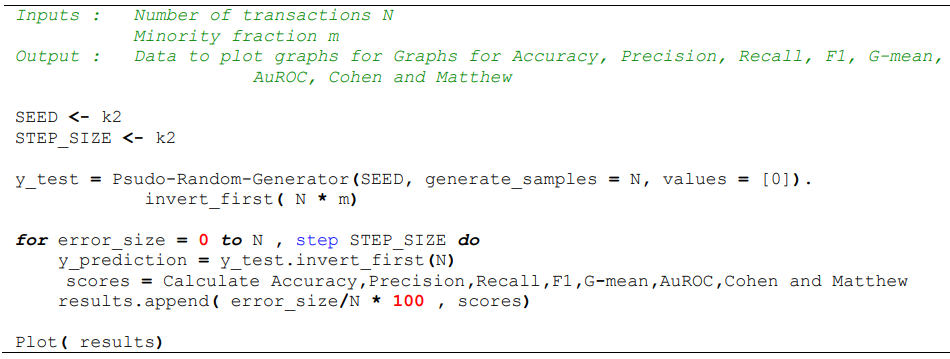}
    \caption{Algorithm for evaluation of scores for different error rates }
    \label{algo_HypotheticalModel}
\end{figure}

There is provision to introduce errors randomly to all classes or to apply it only to the frauds, i.e.: our class of interest, to investigate the effect of noise as well as misclassification errors on the minority class. Figure~\ref{algo_HypotheticalModel} outlines the algorithm of the testing process. 

Next, if we assume source of errors are from the classifier itself, we can now also observe the metrics behavior for \textit{model} introduced classification errors. Thus, we can use our mechanism to study both human annotation errors \textit{and} machine classification errors.


\section{Experiment Results and Analysis}
We select Accuracy, Precision, Recall, F-score, G-mean, AuROC, Cohen Kappa and Mathew as our evaluation metrics from Table~\ref{TABLE_Matrices} and use a SEED of 1234567890 and STEP\_SIZE of 1000.

Using a sample size $N$ of 10,000 we obtained the graphs for minority percentages of 0.5, 0.1, 0.01, 0.001 and 0.0001 for each of the selected evaluation metrics. Note that 0.5 is a perfectly balanced data set. The primary data set has a fraud fraction of 0.0017. 

\subsection{Classifier performance for human annotation error on both classes}
\begin{figure}
    \centering \includegraphics[width=0.9\columnwidth]{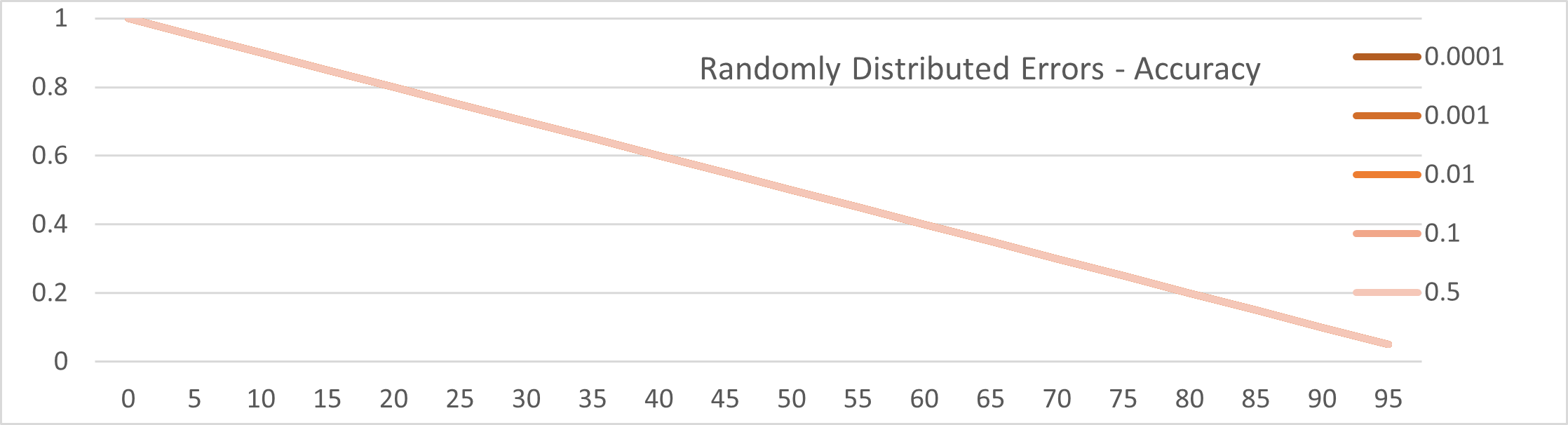}
    \caption{Classifier performance (y-axis) vs Annotation error (x-axis, \textit{both} classes) vs minority fraction (color, [0.0001 \dots 0.5])  for Accuracy metric}
    \label{fig_RandomAcc}
\end{figure}

\begin{figure}
    \centering \includegraphics[width=0.9\columnwidth]{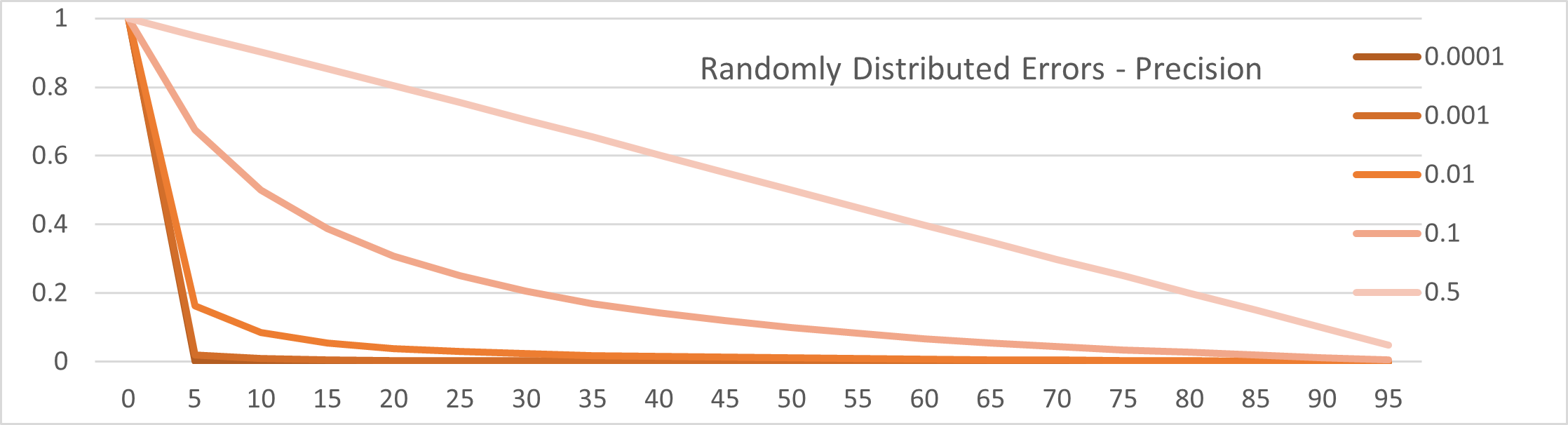}
    \caption{Classifier performance (y-axis) vs Annotation error (x-axis, \textit{both} classes) vs minority fraction (color, [0.0001 \dots 0.5])  for Precision metric}
    \label{fig_RandomP}
\end{figure}

\begin{figure}
    \centering \includegraphics[width=0.9\columnwidth]{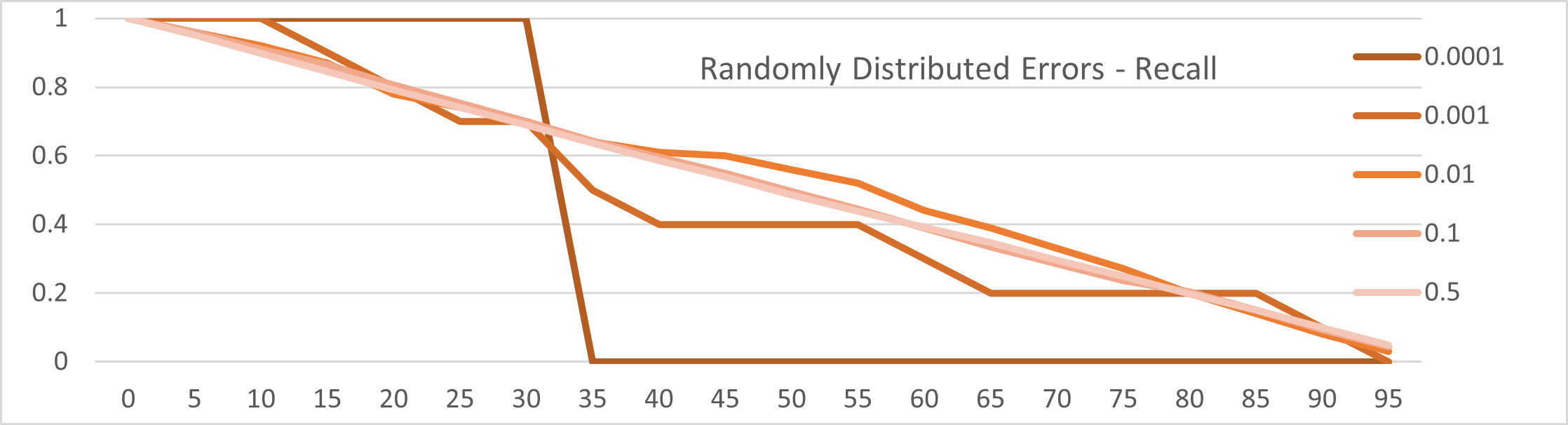}
    \caption{Classifier performance (y-axis) vs Annotation error (x-axis, \textit{both} classes) vs minority fraction (color, [0.0001 \dots 0.5])  for Recall metric}
    \label{fig_RandomR}
\end{figure}

\begin{figure}
    \centering \includegraphics[width=0.9\columnwidth]{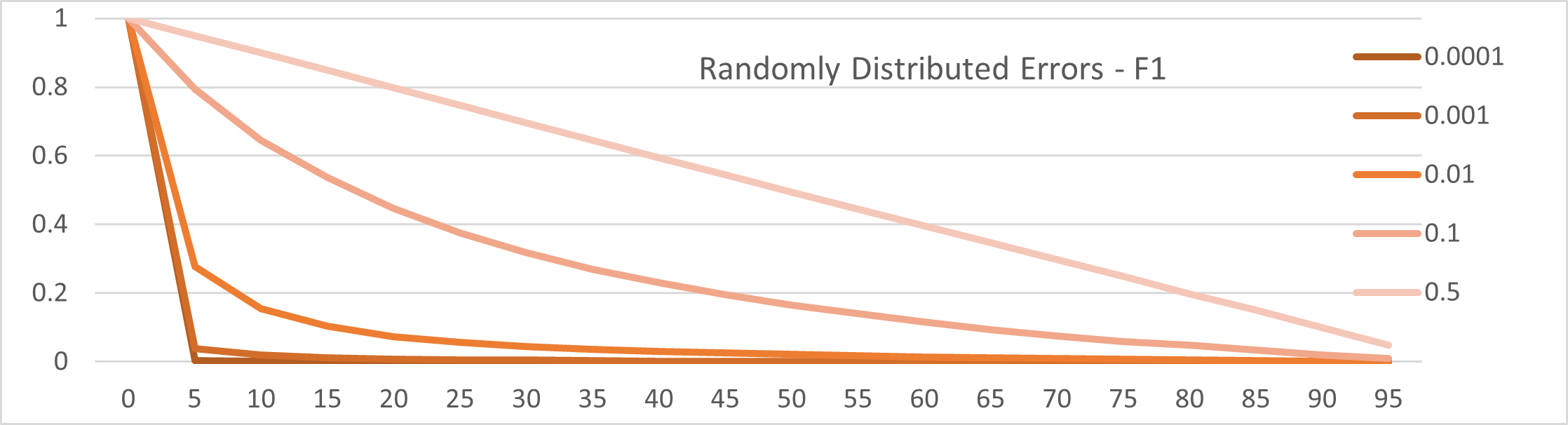}
    \caption{Classifier performance (y-axis) vs Annotation error (x-axis, \textit{both} classes) vs minority fraction (color, [0.0001 \dots 0.5])  for F1 metric}
    \label{fig_RandomF1}
\end{figure}

\begin{figure}
    \centering \includegraphics[width=0.9\columnwidth]{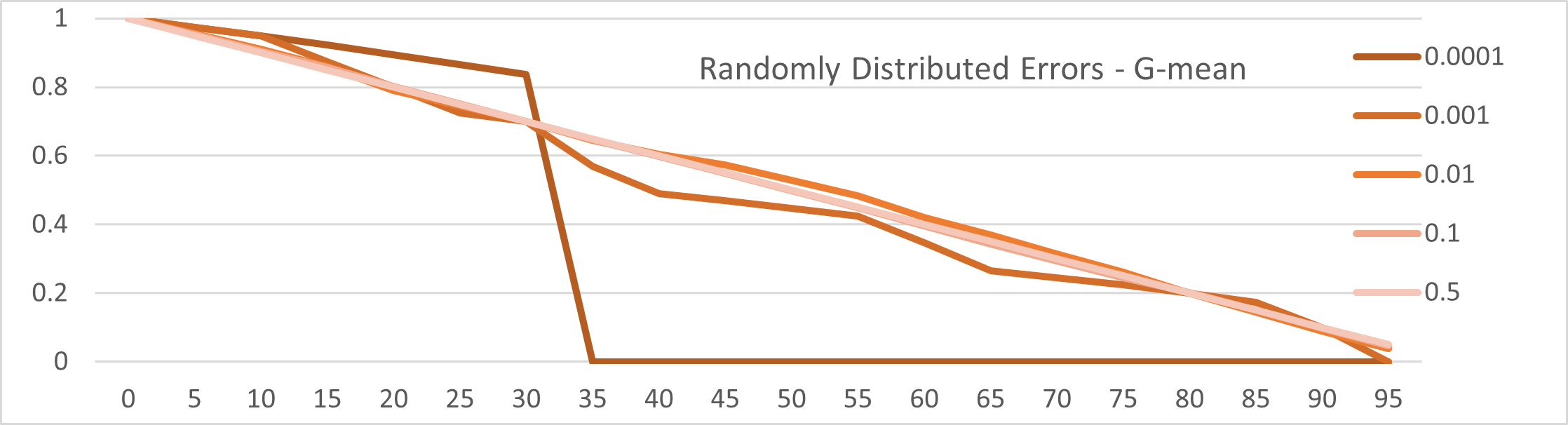}
    \caption{Classifier performance (y-axis) vs Annotation error (x-axis, \textit{both} classes) vs minority fraction (color, [0.0001 \dots 0.5])  for g-mean metric}
    \label{fig_RandomG}
\end{figure}

\begin{figure}
    \centering \includegraphics[width=0.9\columnwidth]{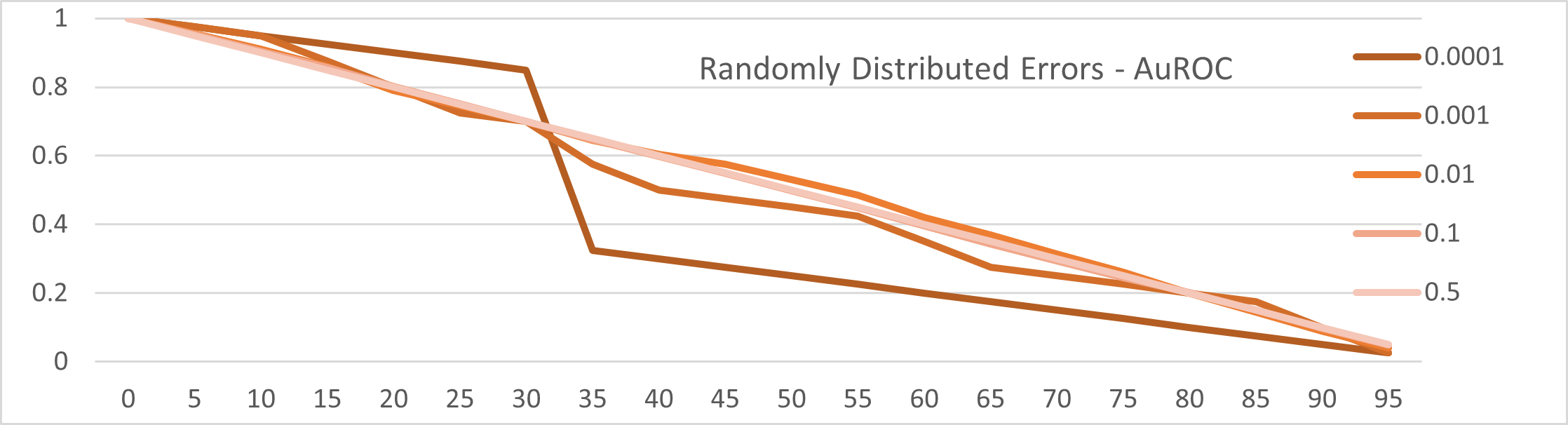}
    \caption{Classifier performance (y-axis) vs Annotation error (x-axis, \textit{both} classes) vs minority fraction (color, [0.0001 \dots 0.5])  for AUC metric}
    \label{fig_RandomAUC}
\end{figure}

\begin{figure}
    \centering \includegraphics[width=0.9\columnwidth]{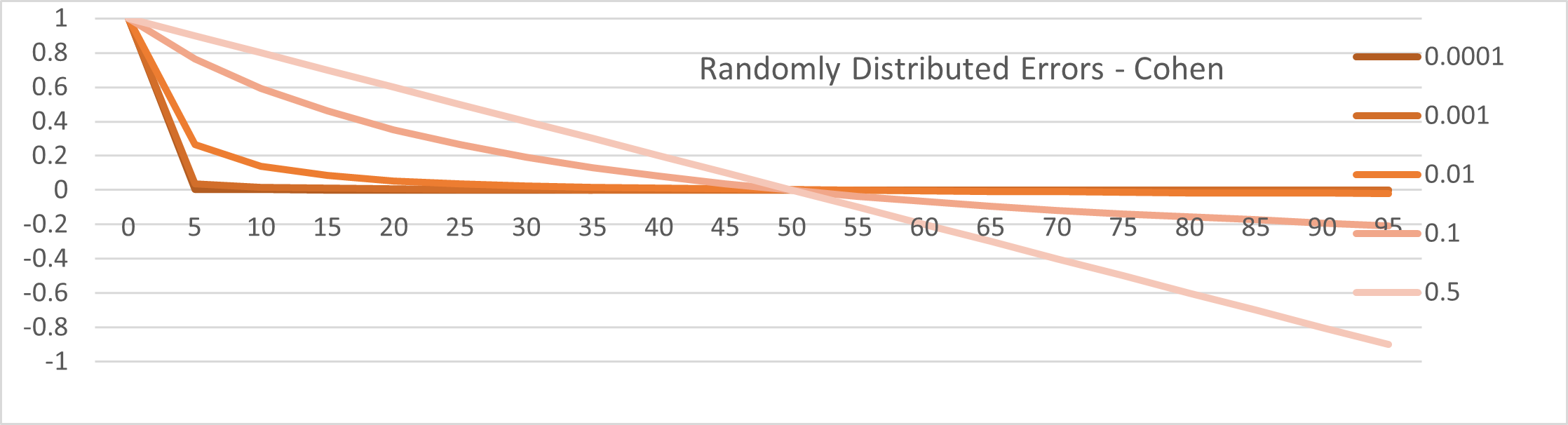}
    \caption{Classifier performance (y-axis) vs Annotation error (x-axis, \textit{both} classes) vs minority fraction (color, [0.0001 \dots 0.5])  for Cohen metric}
    \label{fig_RandomCohen}
\end{figure}

\begin{figure}
    \centering \includegraphics[width=0.9\columnwidth]{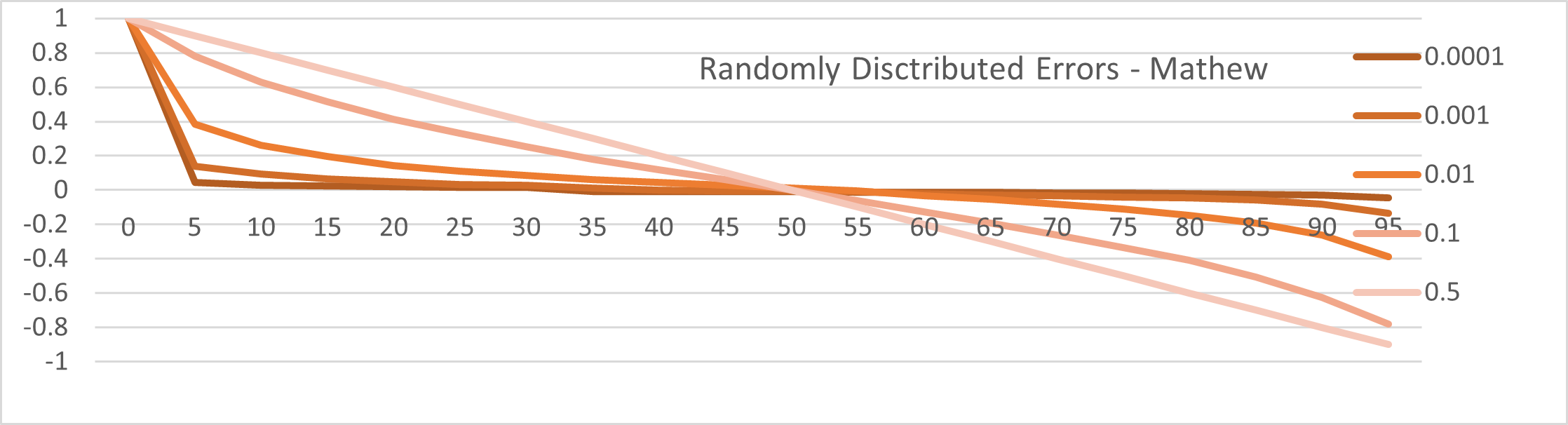}
    \caption{Classifier performance (y-axis) vs Annotation error (x-axis, \textit{both} classes) vs minority fraction (color, [0.0001 \dots 0.5])  for Mathew metric}
    \label{fig_RandomMattew}
\end{figure}

The first set of graphs show randomly distributed errors vs imbalance fraction for each of the evaluation metric. This effectively models classifier performance against imbalance fraction and shows how the metric behaves with increasing imbalance. The classifier errors are randomized and not focused on the class of interest. Figure~\ref{fig_RandomAcc} to Figure~\ref{fig_RandomMattew} shows the relevant graphs.

\subsection{Classifier performance for human annotation error on minority fraud class only}
\begin{figure}
    \centering \includegraphics[width=0.9\columnwidth]{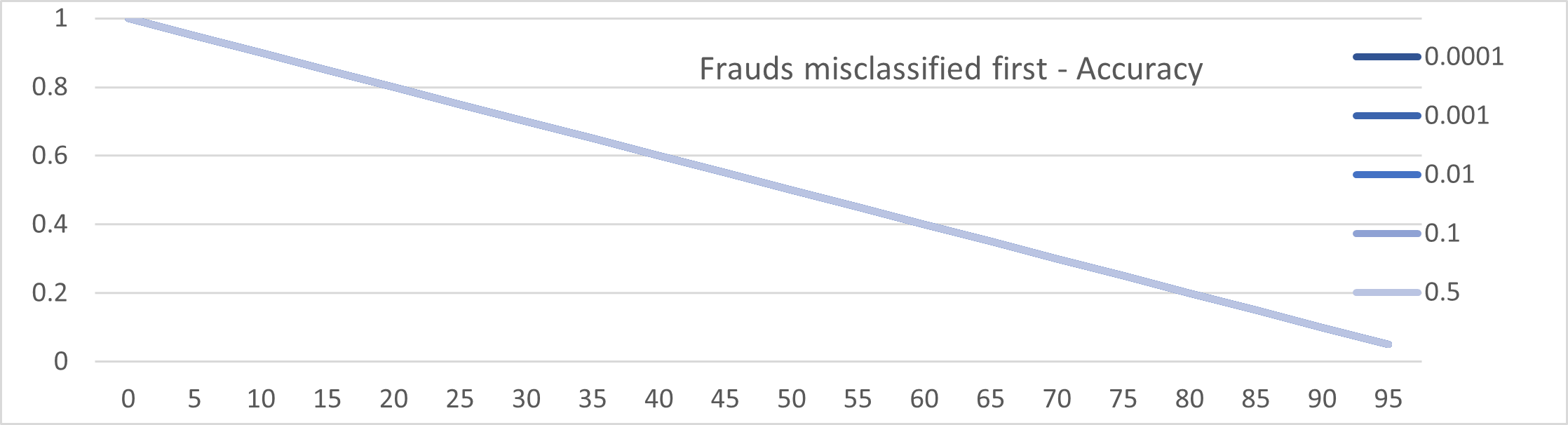}
    \caption{Classifier performance (y-axis) vs Annotation error (x-axis, \textit{minority fraud class only}) vs minority fraction (color, [0.0001 \dots 0.5])  for Accuracy metric}
    \label{fig_FraudAcc}
\end{figure}

\begin{figure}
    \centering \includegraphics[width=0.9\columnwidth]{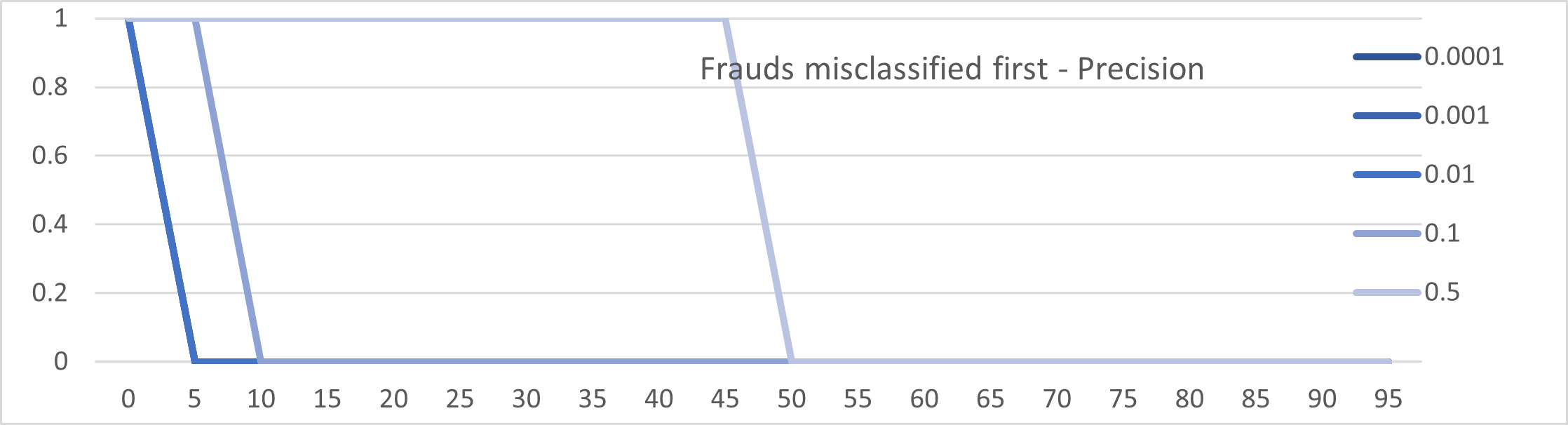}
    \caption{Classifier performance (y-axis) vs Annotation error (x-axis, \textit{minority fraud class only}) vs minority fraction (color, [0.0001 \dots 0.5])  for Precision metric}
    \label{fig_FraudP}
\end{figure}

\begin{figure}
    \centering \includegraphics[width=0.9\columnwidth]{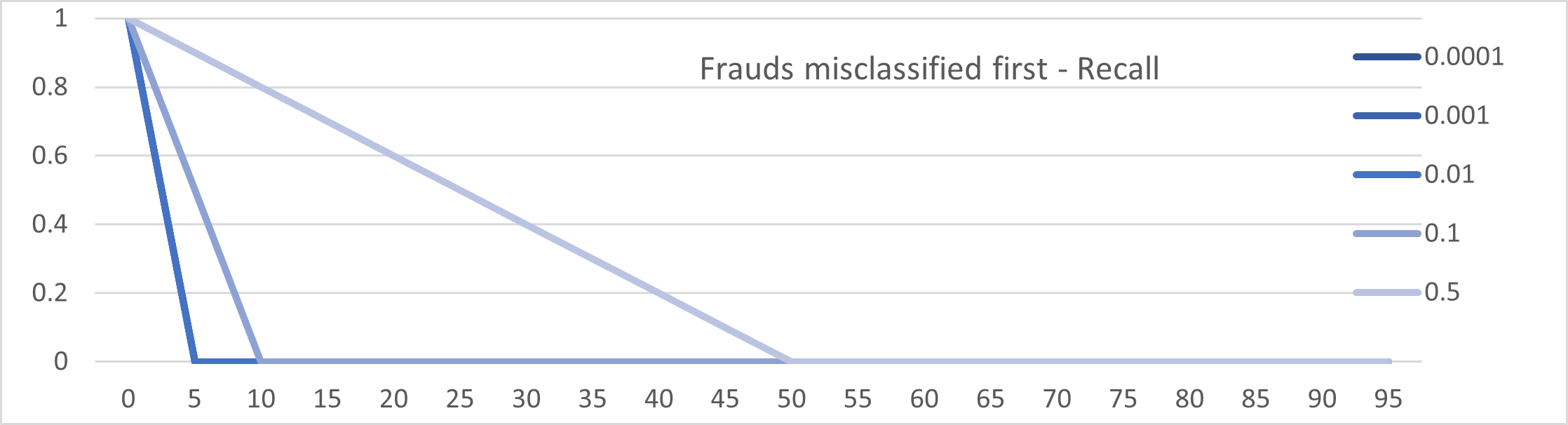}
    \caption{Classifier performance (y-axis) vs Annotation error (x-axis, \textit{minority fraud class only}) vs minority fraction (color, [0.0001 \dots 0.5])  for Recall metric}
    \label{fig_FraudR}
\end{figure}

\begin{figure}
    \centering \includegraphics[width=0.9\columnwidth]{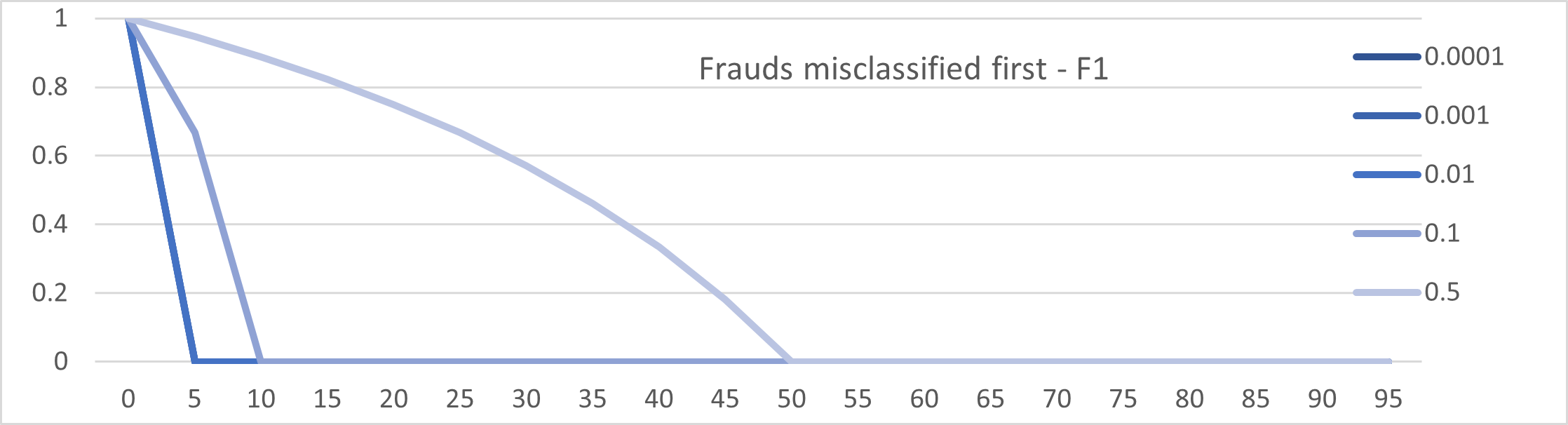}
    \caption{Classifier performance (y-axis) vs Annotation error (x-axis, \textit{minority fraud class only}) vs minority fraction (color, [0.0001 \dots 0.5])  for F1 metric}
    \label{fig_FraudF1}
\end{figure}

\begin{figure}
    \centering \includegraphics[width=0.9\columnwidth]{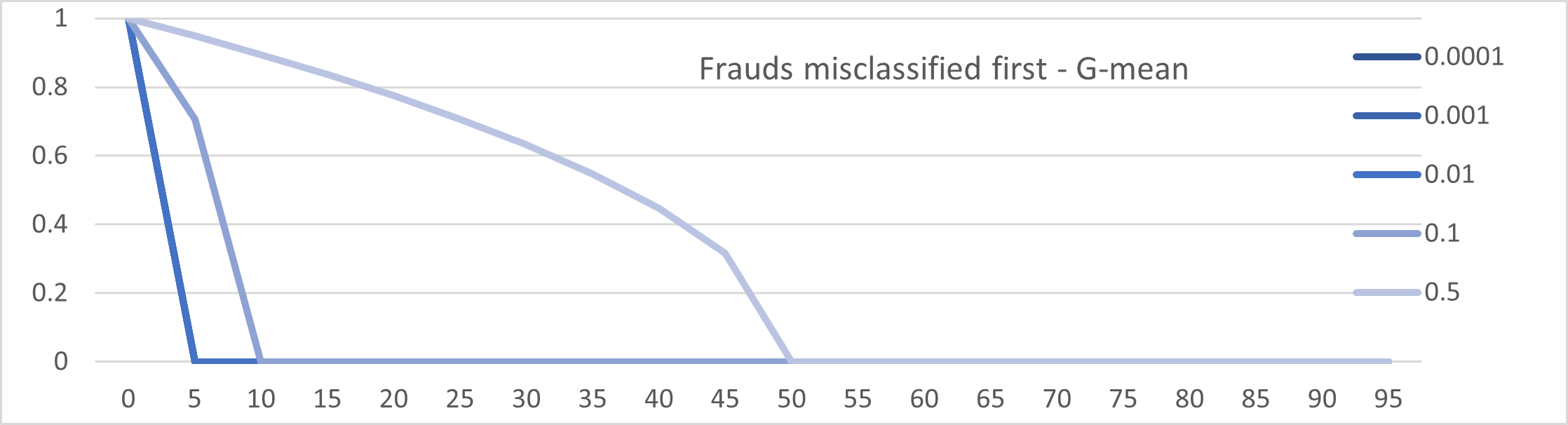}
    \caption{Classifier performance (y-axis) vs Annotation error (x-axis, \textit{minority fraud class only}) vs minority fraction (color, [0.0001 \dots 0.5])  for g-mean metric}
    \label{fig_FraudG}
\end{figure}

\begin{figure}
    \centering \includegraphics[width=0.9\columnwidth]{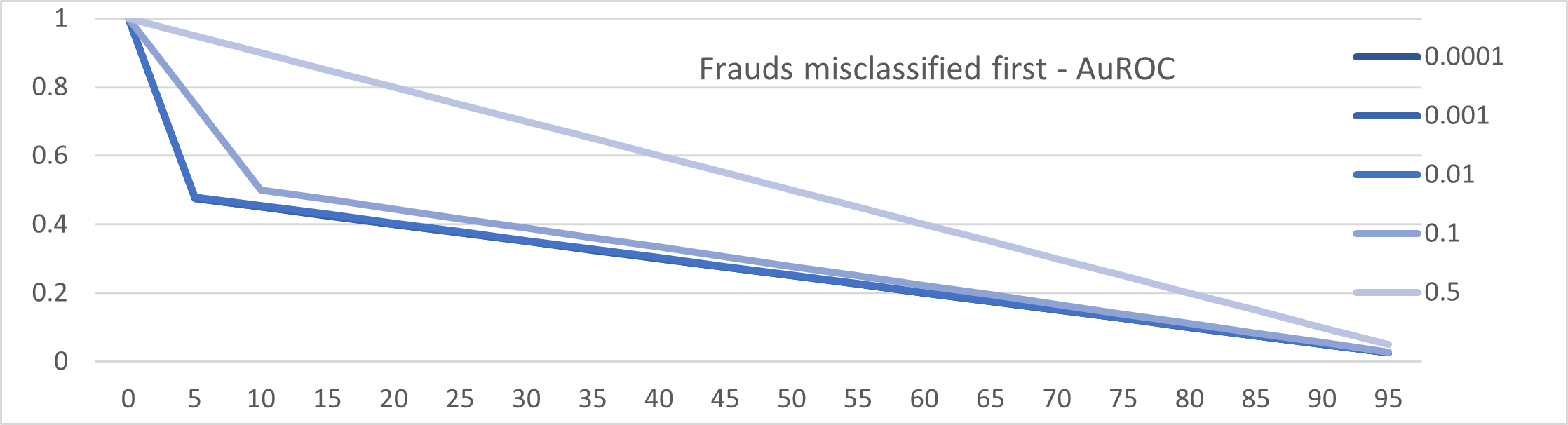}
    \caption{Classifier performance (y-axis) vs Annotation error (x-axis, \textit{minority fraud class only}) vs minority fraction (color, [0.0001 \dots 0.5])  for AUC metric}
    \label{fig_FraudAUC}
\end{figure}

\begin{figure}
    \centering \includegraphics[width=0.9\columnwidth]{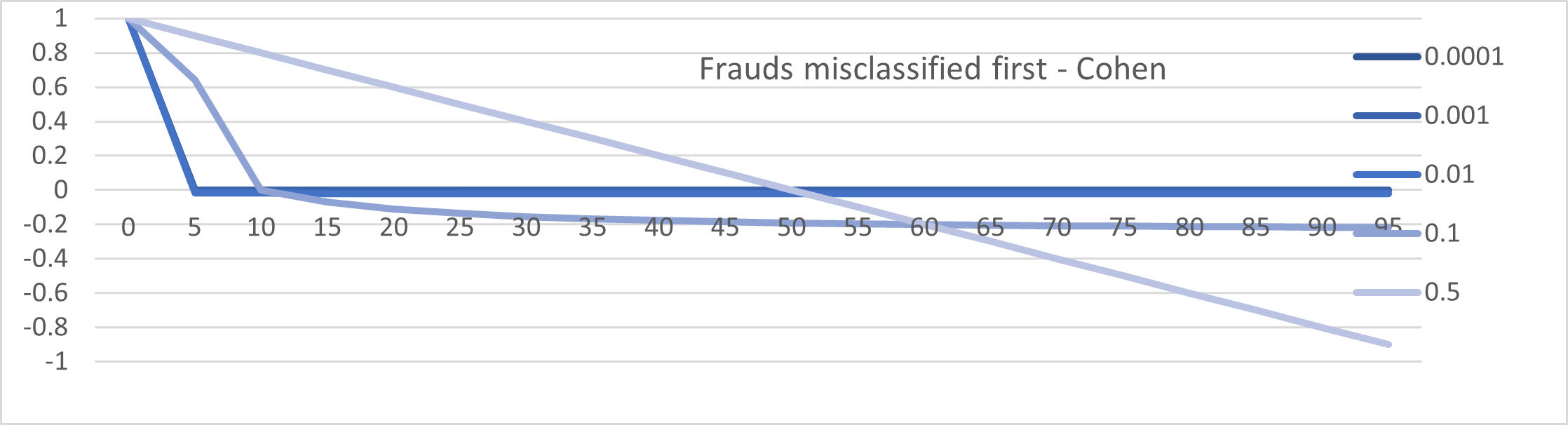}
    \caption{Classifier performance (y-axis) vs Annotation error (x-axis, \textit{minority fraud class only}) vs minority fraction (color, [0.0001 \dots 0.5])  for Cohen metric}
    \label{fig_FraudCohen}
\end{figure}

\begin{figure}
    \centering \includegraphics[width=0.9\columnwidth]{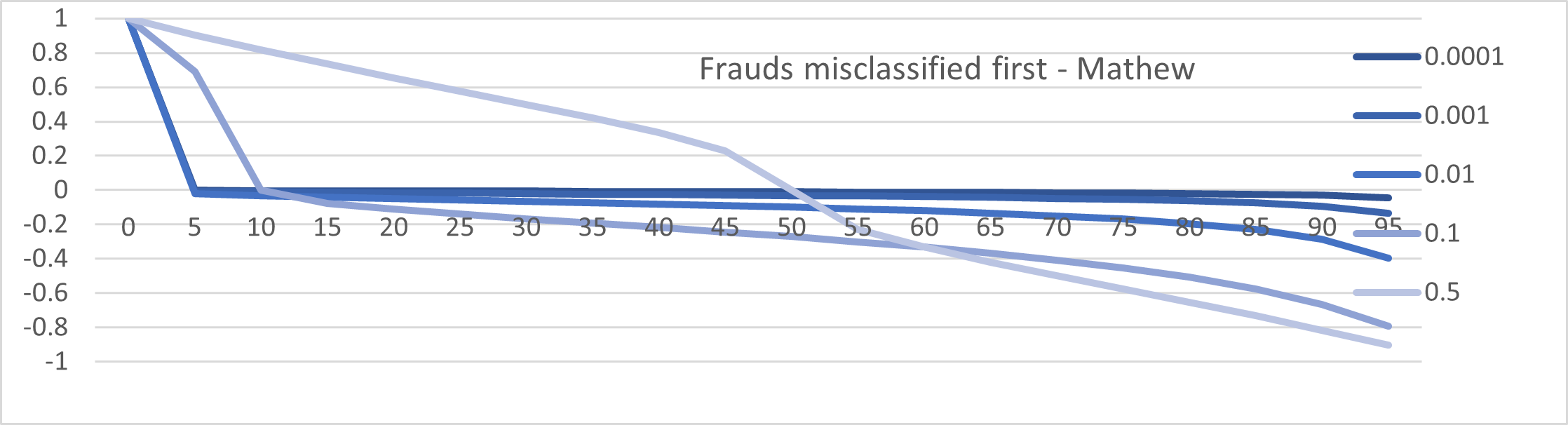}
    \caption{Classifier performance (y-axis) vs Annotation error (x-axis, \textit{minority fraud class only}) vs minority fraction (color, [0.0001 \dots 0.5])  for Mathew metric}
    \label{fig_FraudMattew}
\end{figure}

Figure`\ref{fig_FraudAcc} to Figure~\ref{fig_FraudMattew} shows the behaviour of the evaluation metrics with the misclassification only on the minority fraud class. Since the intention is to identify a suitable evaluation metrics that detects misclassification errors of the frauds, these provide additional insight in to the expected behaviour for classifier scoring. 

\subsection{Analysis}
Accuracy is based on the majority. Therefore, it shows no changes for increase in minority or classification errors on the fraud class. F1 shows a graceful degradation of scoring ability. However, in general, any evaluation metrics do not exhibit any characteristic that makes them better than the others. 

\begin{figure}
    \centering \includegraphics[width=0.8\columnwidth]{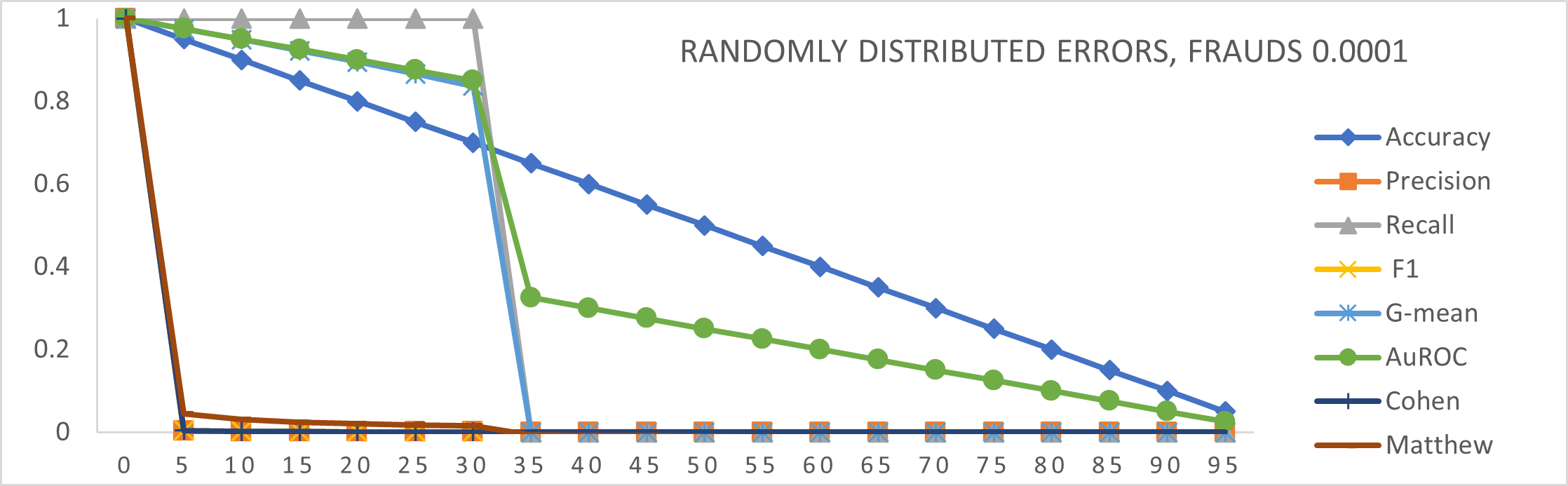}
    \caption{Comparison of Evaluation metrics for randomly distributed annotation error (x-axis, \textit{both} classes) at 0.0001 minority fraction}
    \label{fig_RandomSummary}
\end{figure}

\begin{figure}
    \centering \includegraphics[width=0.8\columnwidth]{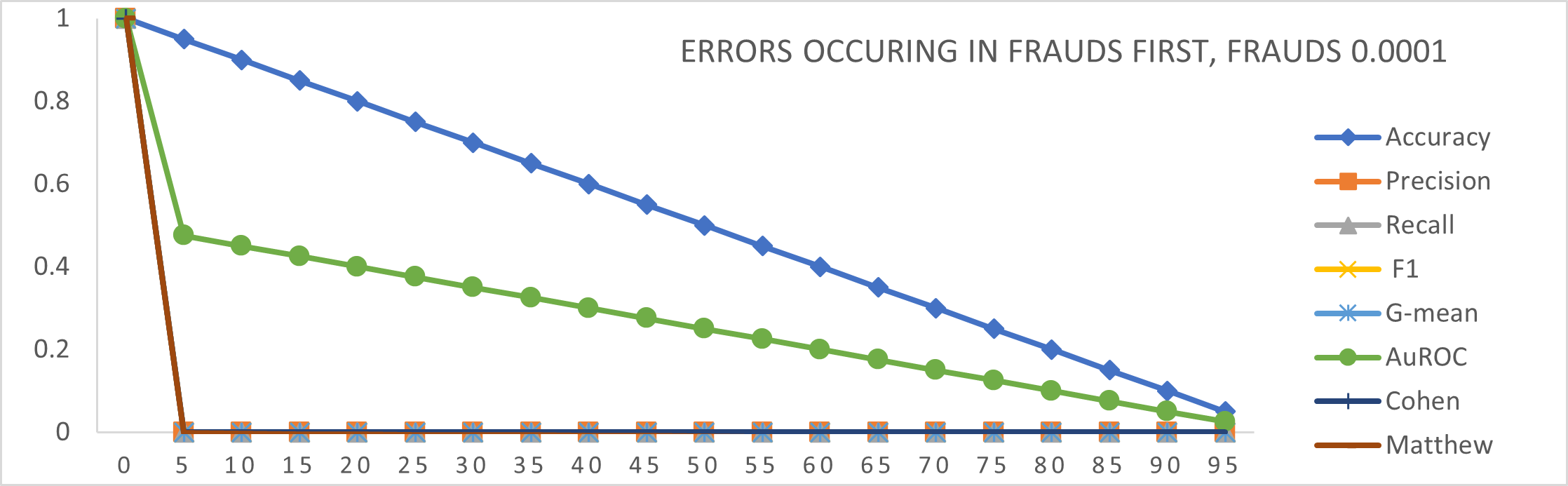}
    \caption{Comparison of Evaluation metrics for randomly distributed annotation error (x-axis, \textit{minority fraud class only}) at 0.0001 minority fraction}
    \label{fig_FraudSummary}
\end{figure}

\begin{figure}
    \centering \includegraphics[width=0.8\columnwidth]{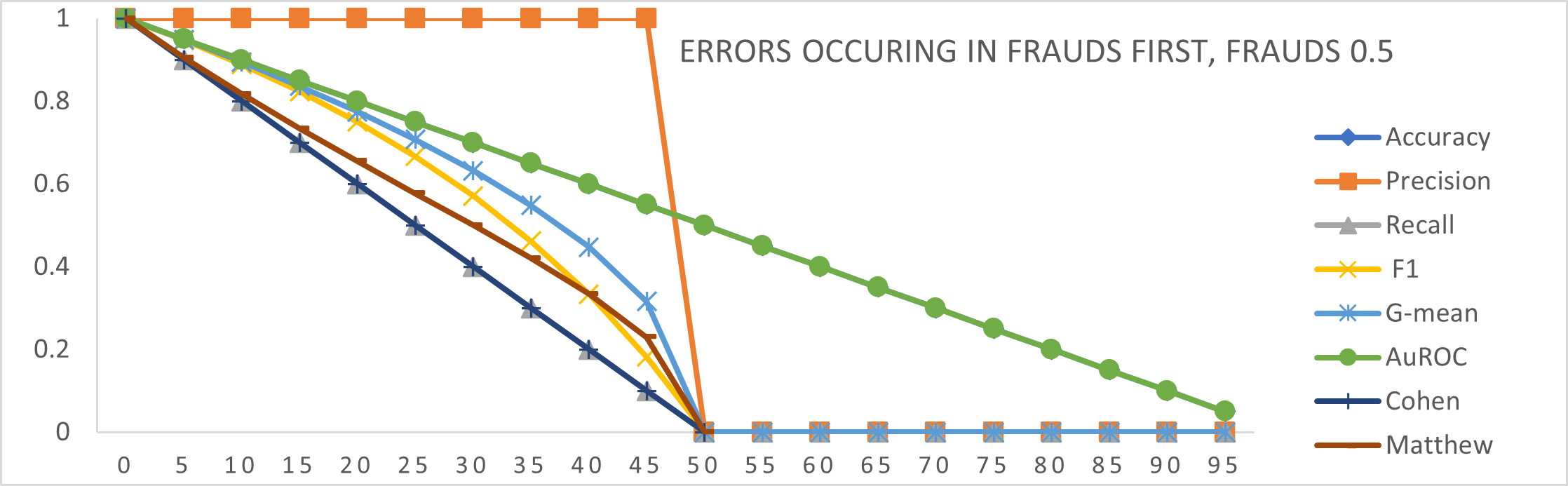}
    \caption{Comparison of Evaluation metrics for randomly distributed annotation error (x-axis, \textit{minority fraud class only}) at 0.5 minority fraction, \textit{the perfectly balanced scenario}}
    \label{fig_BalancedSumamry}
\end{figure}

For completeness and comparison, Figure~\ref{fig_RandomSummary} and Figure~\ref{fig_FraudSummary} shows the combined evaluation metrics behaviors for the extreme imbalanced scenario. Figure~\ref{fig_BalancedSumamry} shows the summary for the evaluation metrics at the (unrealistic and hypothetical), 0.5 perfectly balanced scenario.

While all selected metrics perform well and indicate linearity in the proximity of balanced data set, there is a clear degradation with increasing imbalance. This is more pronounced at the 0.001 range which is, unfortunately, the vital area for fraud detection at the present times. It can be safely assumed that improved fraud detection techniques would push fraud rates even lower and require evaluation metrics that can be effective at increasingly smaller minority fractions in future.

It is possible to gain more insight by focusing on the fraud detection. The evaluation metric should be able to determine the classifier behavior for TP, i.e., identifying the frauds as frauds. It is evident that with even low imbalances, metrics such as Accuracy are unsuitable. However, at extreme 0.001 levels all of the metrics exhibit poor discriminatory power to be able to assess classifier X as n-times better than classifier Y.

It is evident that none of these evaluation metrics present a best or better way of classifier evaluation for characteristic credit card fraud data sets.  

\subsection{Use of a multi-level scoring model}
Given that none of the selected evaluation metrics present a best scoring model, is important to select an \textit{adequate} evaluation metric depending on the specific task a researcher is after. Given the case of massively imbalanced, minority annotation sensitive nature of credit card fraud training data, the following are important in the detection process:
\begin{itemize}
    \item increasing the TP rate (hit rate) to capture all the frauds
    \item keeping the FN (misses) low, to avoid being swamped by normals being misclassified
\end{itemize}

F1 score, is the harmonic mean of precision and recall. Recall, or Sensitivity measures how often did the classifier manage to detect a fraud while, Precision, or the Hit rate measures the accuracy of how often the detection was actually correct. Unfortunately, F1 favors models that have high Precision and Recall. For instances where one would prefer a high Precision (predictions you make are accurate) or high Recall (where the classifier makes more predictions) over the other F1 tends to be small. This is due to the fact that increasing Precision generally lowers Recall, which is known and the precision/recall trade-off.

The general $F_\beta$ score is given by:
\begin{equation} 
    F_\beta=(1+\ \beta^2)\cdot \frac{precision\cdot recall}{(\beta^2\cdot precision)+recall}
\label{eq:FBeta}
\end{equation}
where, $\beta = 1$ makes Precision and Recall equally important, while $\beta > 1$ means more Precision and lesser Recall. Unfortunately, F1 (as well as Recall and Precision) is biased to the majority normals class.

G-mean, on the other hand, measures the balance between the fraud and normal classes. G-mean attempts to maximize the accuracy on each class in a binary classification problem \cite{barandela2003strategies}. 
It is ideal to avoid overfitting in the majority normals class and underfitting in the minority fraud class. Importantly, even if normals (negatives) are well classified, g-mean would still give a low value if frauds (positives) are classified poorly. According to \cite{al2016algorithms}, g-mean is the most acceptable evaluation matric for imbalanced data.

Therefore, based on the findings of this section, the composite evaluation metrics, comprising of F1 and g-mean scoring, in that given order, is proposed to be used to evaluate massively imbalanced credit card fraud data. Further, as seen in the PCA section, F1 and g-mean act as independent scores from different perspectives for massively unbalanced credit card fraud data sets.

\section{Conclusion}

In this work, we develop a theoretical foundation to model human annotation errors and extreme imbalance typical in real world fraud detection data sets. 
We conduct empirical experiments with a synthetic data distribution, approximated to a popular real world credit card fraud data set, to simulate human annotation errors and extreme imbalance to observe the behavior of popular machine learning classifier evaluation matrices. 
Specifically, we investigate the behavior of Accuracy, Precision, Recall, F1, G-mean, AuROC, Cohen, Matthew scores in the presence of massive imbalance and increasing human annotation errors; for both the general case where all classes are affected as well as when only the more important (minority) fraud class is affected. 
We demonstrate that a combined F1 score and g-mean, in that specific order, is the best evaluation metric for massively imbalanced, minority annotation sensitive data sets such as fraud detection.
In the process, a hypothetical classifier-based evaluation mechanism is proposed, developed and made available to the research community for future evaluation metrics assessment that is independent of an underlying model. This is able to simulate classification errors, imbalance and noise. 

In future work, we hope to explore the development of more robust and universal evaluation metrics for fractional percentage minority class detection capable machine learning model evaluation. This would open up and unify the model development and benchmarking for typical extremely rare nevertheless significant scenarios such as earthquakes, tsunamis, hurricanes, floods, asteroid impacts, solar flares, bank failures, market crashes etc.

\section{Acknowledgments}
Dedicated to beloved parents.

\bibliographystyle{splncs04}
\bibliography{main}

\begin{thebibliography}{10}
\providecommand{\url}[1]{\texttt{#1}}
\providecommand{\urlprefix}{URL }
\providecommand{\doi}[1]{https://doi.org/#1}

\bibitem{al2016algorithms}
Al~Helal, M., Haydar, M.S., Mostafa, S.A.M.: Algorithms efficiency measurement
  on imbalanced data using geometric mean and cross validation. In: 2016
  international workshop on computational intelligence (IWCI). pp. 110--114.
  IEEE (2016)

\bibitem{barandela2003strategies}
Barandela, R., Sanchez, J., Garcia, V., Rangel, E.: Strategies for learning in
  class imbalance problems. Pattern Recognition  \textbf{3}(36),  849--851
  (2003)

\bibitem{dal2017credit}
Dal~Pozzolo, A., Boracchi, G., Caelen, O., Alippi, C., Bontempi, G.: Credit
  card fraud detection: a realistic modeling and a novel learning strategy.
  IEEE transactions on neural networks and learning systems  \textbf{29}(8),
  3784--3797 (2017)

\bibitem{dal2015calibrating}
Dal~Pozzolo, A., Caelen, O., Johnson, R.A., Bontempi, G.: Calibrating
  probability with undersampling for unbalanced classification. In: 2015 IEEE
  symposium series on computational intelligence. pp. 159--166. IEEE (2015)

\bibitem{duman2013solving}
Duman, E., Elikucuk, I.: Solving credit card fraud detection problem by the new
  metaheuristics migrating birds optimization. In: International
  Work-Conference on Artificial Neural Networks. pp. 62--71. Springer (2013)

\bibitem{ebert2016machine}
Ebert, C., Louridas, P.: Machine learning. IEEE Software  \textbf{33}(5),
  110--115 (2016)

\bibitem{kulatilleke2022Challenges}
Kulatilleke, G.K.: Challenges and complexities in machine learning based credit
  card fraud detection. arXiv preprint arXiv:2208.10943  (2022)

\bibitem{kulatilleke2022Credit}
Kulatilleke, G.K.: Credit card fraud detection - classifier selection strategy.
  arXiv preprint arXiv:  (2022)

\bibitem{lichman2013uci}
Lichman, M., et~al.: Uci machine learning repository (2013)

\bibitem{makki2017fraud}
Makki, S., Haque, R., Taher, Y., Assaghir, Z., Ditzler, G., Hacid, M.S.,
  Zeineddine, H.: Fraud analysis approaches in the age of big data-a review of
  state of the art. In: 2017 IEEE 2nd international workshops on foundations
  and applications of self* systems (FAS* W). pp. 243--250. IEEE (2017)

\bibitem{sahin2013cost}
Sahin, Y., Bulkan, S., Duman, E.: A cost-sensitive decision tree approach for
  fraud detection. Expert Systems with Applications  \textbf{40}(15),
  5916--5923 (2013)

\bibitem{sathyapriya2017big}
Sathyapriya, M., Thiagarasu, V.: Big data analytics techniques for credit card
  fraud detection: A review. International Journal of Science and Research
  \textbf{6}(5),  206--211 (2017)

\bibitem{west2016some}
West, J., Bhattacharya, M.: Some experimental issues in financial fraud mining.
  Procedia Computer Science  \textbf{80},  1734--1744 (2016)

\bibitem{zareapoor2015application}
Zareapoor, M., Shamsolmoali, P., et~al.: Application of credit card fraud
  detection: Based on bagging ensemble classifier. Procedia computer science
  \textbf{48}(2015),  679--685 (2015)

\bibitem{zojaji2016survey}
Zojaji, Z., Atani, R.E., Monadjemi, A.H., et~al.: A survey of credit card fraud
  detection techniques: data and technique oriented perspective. arXiv preprint
  arXiv:1611.06439  (2016)

\end{thebibliography}
\end{document}